\documentclass{article}
\usepackage{amsmath,graphicx,mlspconf}   
\usepackage{url}            
\usepackage{booktabs}       
\usepackage{amsfonts}       
\usepackage{nicefrac}       
\usepackage{microtype}      
\usepackage{bm}
\usepackage{amsmath}
\usepackage{amsfonts}
\usepackage{color}
\usepackage[makeroom]{cancel}
\usepackage{cancel}
\usepackage{tabu}
\usepackage[normalem]{ulem} 
\usepackage{framed}
\usepackage{algorithm}
\usepackage{algorithmic}
\usepackage[draft,bookmarks=false]{hyperref} 

\def\0{{\mathbf 0}}

\newcommand{\xc}{\bm{x}}

\newcommand{\zc}{\bm{z}}

\copyrightnotice{978-1-7281-6662-9/20/\$31.00 {\copyright}2020 IEEE}

\toappear{2020 IEEE International Workshop on Machine Learning for Signal Processing, Sept.\ 21--24, 2020, Espoo, Finland}

\title{Multinomial Sampling for Hierarchical Change-Point Detection}
\name{%
    Lorena Romero-Medrano$^{*}$%
    \qquad Pablo Moreno-Mu\~noz$^{*}$%
    \qquad Antonio Art\'es-Rodr\'iguez%
    \thanks{$^{*}$ Equal contribution. This work was supported by the Ministerio de Ciencia, Innovaci{\'o}n y Universidades under grant TEC2017-92552-EXP (aMBITION), by the Ministerio de Ciencia, Innovaci{\'o}n y Universidades, jointly with the European Commission (ERDF), under grants TEC2017-86921-C2-2-R (CAIMAN) and RTI2018-099655-B-I00 (CLARA), and by the Comunidad de Madrid under grant Y2018/TCS-4705 (PRACTICO-CM). The work of PMM has been supported by FPI grant BES-2016-077626 and LRM has been also supported by grant IND2018/TIC-9649 from the Comunidad de Madrid.}
}
\address{%
    Dept. of Signal Theory and Communications, Universidad Carlos III de Madrid, Spain \\%
    Gregorio Mara\~n\'on Health Research Institute, Spain%
}

\begin{document}

\maketitle

\begin{abstract}
Bayesian change-point detection, together with latent variable models, allows to perform segmentation over high-dimensional time-series. 
We assume that change-points lie on a lower-dimensional manifold where we aim to infer subsets of discrete latent variables. 
For this model, full inference is computationally unfeasible and pseudo-observations based on point-estimates are used instead. 
However, if estimation is not certain enough, change-point detection gets affected.
To circumvent this problem, we propose a multinomial sampling methodology that improves the detection rate and reduces the delay while keeping complexity stable and inference analytically tractable.
Our experiments show results that outperform the baseline method and we also provide an example oriented to a human behavior study. 

\end{abstract}
\section{Introduction}
\label{sec:intro}

Change-point detection (CPD) aims to identify abrupt transitions in sequences of observations, for both univariate or multivariate cases. Typically, a change-point is only considered if there is a noticeable difference between the generative parameters of data. Two classical families of approaches are identified in signal processing and machine learning. First, the main focus of early literature has been on \textit{batch} settings \cite{scott1974cluster,fearnhead2006exact}, where the entire data set is available for processing. Second, \textit{online} CPD methods \cite{Adams2007} avoid the previous assumption to fulfill two intertwined tasks: i) estimation of the generative model parameters as observations come in and ii) segmentation of data into partitions based on the parameters obtained.

The identifiability of change-points (CP) is directly related to the discrepancy between the distributions governing each partition.
In this context, the Bayesian framework provides a solution to obtain uncertainty measures over both parameters and CP positions. The Bayesian online CPD algorithm (BOCPD) of \cite{Adams2007} uses this idea to derive a recursive exact inference method.
However, when observations become high-dimensional and parameters grow exponentially, there is not enough evidence on the sequential data to obtain reliable estimates of the generative parameters.

Latent variable models are amenable to overcome the high-dimensionality issue.  Under the assumption that change-points lie on a lower-dimensional manifold, one can extend the BOCPD algorithm to accept surrogate mixture models \cite{MorenoRamirezArtes18}. The main drawback is that true latent class assignments are never observed but inferred, leading to introduce pseudo-observations. For this purpose, there are two main strategies: i) use the posterior probability vector as a continuous multivariate datum or ii) observe single point-estimates of the latent variable. While the first idea requires expensive approximate methods due to non-tractability, the second one allows reliable detection when posterior distributions over the latent variables are certain enough.

In this paper, we consider the case of having poor inference point-estimates over the latent variables that lead to catastrophic results on the CPD. Our contribution is to provide a novel extension for the hierarchical model that improves the detection rate and delay even under extremely \textit{flat} posterior distributions.  The solution considers latent variable samples as multivariate observations that are multinomial distributed. It keeps the original analytic simplicity of inference as well as the complexity cost remains low. %
In the experiments, we prove the utility of the new inference method on synthetic data and we also provide insights to be applicable in real-world scenarios, such as change-point detection in a human behavior study.
%




\section{Bayesian Change-Point Detection}
\label{sec:bocpd}
We assume that a sequence of observations $\xc_{1}, \xc_{2}, \dots, \xc_{t}$ may be partitioned into non-overlapping segments. Each segment or partition $\rho$ with $\rho=\{1,2,\dots\}$ has a surrogate generative distribution $p(\xc|\bm{\theta}_{\rho})$ where parameters $\bm{\theta}_{\rho}$ are unknown and observations are independent and identically distributed (i.i.d). The maximum number of partitions is also unknown and unbounded, and it may increase as a new datum $\xc_t$ comes in. 

Based in \cite{Adams2007}, we are concerned with discovering the true generative distributions $p(\xc|\bm{\theta}_{t})$ and hence, their parameters $\bm{\theta}_{t}$ at each time-step. To alleviate the combinatorial problem of estimating parameters based on every partition hypothesis, we introduce an auxiliary random variable (r.v.) $r_t$, also called the \textit{run-length} in the original version of \cite{Adams2007}. The discrete variable counts the number of time-steps since the last change-point, that is
\begin{equation}
r_t = \begin{cases} 0, & \textsc{cp} \text{ at time } t \\
r_{t-1} + 1, & \text{otherwise}. \\
\end{cases}
\end{equation}
The main idea behind the $\textit{run-length}$, $r_t$, is that it converts the partition hypothesis problem into a Bayesian inference task as well as serves as a relatively simple CP indicator. This strategy augments the model, leading to a double inference mission: i) estimating  the posterior distribution over $r_t$ and ii) obtaining reliable values of $\bm{\theta}_{t}$ parameters. 

\begin{figure}[t!]
	\centering
	\includegraphics[width=0.85\columnwidth]{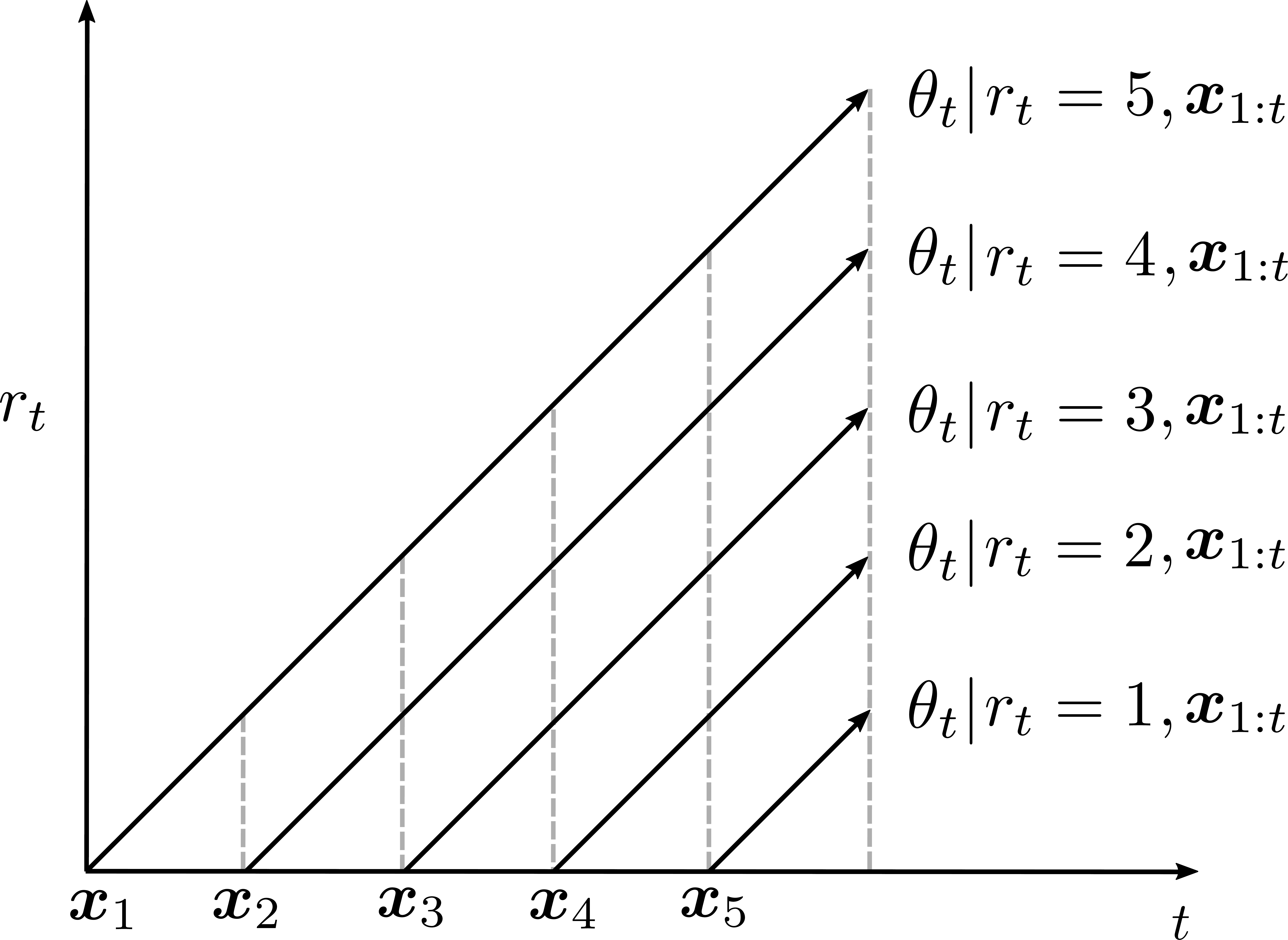}
	\caption{Illustration of the parallel inference mechanism for the estimation of $\bm{\theta}_t$ conditioned on the run-length $r_t$ given $\xc_{1:t}$.}
	\label{fig:thread}
\end{figure}

The discrete nature of the $r_t$ counting r.v.\ also makes it appropriate for integration, being feasible to obtain the posterior distribution $p(r_t|\xc_{1:t})$ in a recursive manner. Importantly, the factorization of the joint distribution $p(r_t,\xc_{1:t})$ presented in \cite{Adams2007} is based on the marginalization of model parameters $\bm{\theta}_t$. When this is not possible, for instance, due to the underlying generative model $p(\xc|\bm{\theta}_{t})$ is more expressive or complex, other ways for approximate inference must be considered \cite{Saatci2010,Turner2013}.

The learning of $\bm{\theta}_t$ is conditioned to the run-length $r_t$ and hence, the partition hypothesis, carrying out a multiple-thread inference mechanism. For example, having observed $\xc_{1:5}$ at some time-step $t=5$, we would have to compute several posterior estimations $\bm{\theta}_t|r_t, \xc_{1:t}$, one per each $r_t$ value. As a consequence, given $r_t = 2$, the estimation would be analogous to $\bm{\theta}_t|\{\xc_4, \xc_5\}$ under the previous notation. This example is depicted in the graphical scheme of Figure 1.

However, a key inconvenient appears as the size of observations $\xc_t$ rises, and the method works in a potentially high-dimensional setting. In such cases, the complexity of the generative model increases accordingly, leading to an extremely large number of parameters $\bm{\theta}_t$ to estimate. This fact makes almost impossible to perform CPD in a reliable manner as there is not sufficient statistical evidence given $\xc_{1:t}$, to update our posterior distribution. In such case, CPs are typically confounded with noise drifts in the underlying parameters.

\section{CPD and Latent Variable Models}

Latent variable models are a powerful tool in unsupervised learning, with significant connections with Bayesian statistics. This family of approaches typically assumes that there exists a finite low-dimensional representation of the data that characterizes the generative properties of observed objects. When the high-dimensionality problem appears, it has become a popular solution in probabilistic modelling as it allows to easily take decisions about the dimensionality of the latent manifold, its nature (i.e.\ continuous or discrete) and its conditioning with the rest of r.v.\ implied in the generative model of data. 

In our particular case for Bayesian CPD, we may assume that the observations $\xc_{1:t}$ belong to a lower-dimensional manifold, where the true CPs lie. The generative model is then expressed as
\begin{equation}
	p(\xc_{t}|\bm{\theta}_t) = \int p(\xc_t|z_t)p(z_{t}|\bm{\theta}_t)dz_t,
\end{equation}
where the conditional distribution $p(\xc_t|z_t)$ is assumed to be fixed and $p(z_{t}|\bm{\theta}_t)$ is the new likelihood distribution over the latent variable $z_t$, that can be either continuous or discrete. Similar ideas were previously explored in \cite{MorenoRamirezArtes18,AguGomBauSchPet20} as extensions of the BOCPD method, where only discrete $z_t$ variables were considered.

\subsection{Hierarchical CPD}
We introduce the hierarchical model in \cite{MorenoRamirezArtes18}, where $z_t$ is a categorical r.v.\ or \textit{class}, such that $z_t = \{1,2,\dots,K\}$, working as the assignment of each observation object $\xc_t$. In the CPD scenario, it can be understood as a segmentation problem of different mixture models. As long as we cannot observe the true assignments $z_t$, we instead use \textit{maximum a-posteriori} (MAP) estimates as our pseudo-observations. The point-estimates are obtained from
\begin{equation}
\label{eq:argmax}
z_t^{\star} = \arg \max_{z_t} p(z_t|\xc_t),
\end{equation}
where we have previously performed inference to obtain $p(z_t|\xc_t)$. For instance, via the online expectation-maximization (EM) algorithm \cite{cappe2009online} or other continual learning strategies \cite{MorenoRamirezArtes20}.

The use of sequences of MAP estimates $\zc^{\star}_{1:t}$ allows us to build the following joint distribution
\begin{equation}
\label{eq:recursive}
p(r_t,\zc^{\star}_{1:t}) = \sum_{r_{t-1}} p(r_t|r_{t-1})\Psi^{(r)}_tp(r_{t-1}, \zc_{1:t-1}^{\star}),
\end{equation}
where $p(r_t|r_{t-1})$ is the conditional prior that modulates how likely is to detect a new CP, that is $r_t=0$, given the previous run-length hypothesis $r_{t-1}$. The full development of the joint probability and the details of $p(r_t|r_{t-1})$ are explained in \cite{MorenoRamirezArtes18}. The predictive term $\Psi^{(r)}_t$ is obtained as
\begin{equation}
\Psi^{(r)}_t = \int p(z_t^{\star}|\bm{\theta}_t)p(\bm{\theta}_t|r_{t-1}, \zc_{1:t-1}^{\star}) d\bm{\theta}_t, \label{eq:predictive}
\end{equation}
with $p(\bm{\theta}_t|r_{t-1}, \zc_{1:t-1}^{\star})$ being the multiple posterior updates depicted in the diagram of Figure \ref{fig:thread}. However, working with the sequence of MAP point-estimates may lead to false-alarm or missing detection problems when the inferred posterior distribution $p(\zc_{1:t}|\xc_{1:t})$ is extremely \textit{flat}. If the MAP estimation does not coincide with the true latent class assignment, it would introduce noise in the CPD with undesired results.


\section{Multinomial Sampling}
\label{sec:multinomial}

Our goal is to obtain a better characterization of the underlying posterior distribution at each time step $t$, when this is not well fitted. We can generate \textit{pseudo-observations} of the latent variable by drawing $S$ i.i.d. samples of the posterior distribution $z^{(1)}_t, z^{(2)}_t, \dots, z^{(S)}_t \sim p(z_{t}| \xc_{t})$ $\forall t$,
rather than working with a single point-estimate $z^{\star}_t$. 

The new approach addresses the question of how to deal with a subset of $S$ samples instead of one at each time step. A potential idea would be to introduce Monte-Carlo (MC) sampling methods, but it would lead to draw $S\cdot t$ samples at each time step, becoming unfeasible in the long term. Alternatively, we propose to assume samples multinomial-distributed, which preserves the prior-conjugacy and is still consistent with the BOCPD algorithm presented in \cite{Adams2007}. 

A multinomial distribution with parameters $\bm{\theta}_t\in\mathcal{S}^K$ and $N,$ measures the probability that  each class $k\in\{1,...,K\}$ has been observed $n_k$ times over $N$ categorical independent trials with same probabilities $\bm{\theta}_t$. 
This model allows us to deal with an augmented number of observations at each time $t$ with just the cost of introducing one more parameter in the model: $N=S$, the total number of samples drawn from the posterior.

Given the sampled vector $\bm{z}_t^{\star} = (z_t^{(1)}, z_t^{(2)}, \dots, z_t^{(S)})\in\{1,...,K\}^S$, we can define its associated counting vector $\bm{c}_t \in\mathbb{Z}_+^K$ where $c_t^{k}:= \sum_{s=1}^S\mathbb{I}\{z_t^{(s)}=k\}$ $\forall k$, therefore having $\sum_{k=1}^Kc_t^k = S$.
Thus, at each time $t$, we can consider the counting vector $\bm{c}_t$ as an i.i.d.\ observation of a multinomial distribution with natural parameters $\bm{\theta}\in\mathcal{S}^K$ and $ S\in\mathbb{N}$. 

With the previous notation and assuming that
\begin{equation}
\begin{split}
\bm{\theta}_t \sim \text{Dirichlet}(\bm{\alpha}),\\
\bm{c}_t\sim \text{Multinomial}(\bm{\theta}_t, S),
\end{split}
\end{equation}
where $\bm{\alpha}\in\mathbb{R}_+^K$ and the likelihood expression of $\bm{c}_t$ is
\begin{equation}
\label{likelihood}
p(c_t^1,...,c_t^K| \bm{\theta}, S) = \frac{S!}{\prod_{k=1}^Kc_t^k!}\prod_{k=1}^K\theta_{k}^{c_t^k}.
\end{equation}
The posterior update of parameters has the following closed form $\bm{\alpha}' = \bm{\alpha} + \bm{c}_t$, allowing a direct update of the parameters when a new sample is observed.

\begin{algorithm}[ht!]
	\caption{Multinomial CPD }
	\label{alg:simplified}	
	\begin{algorithmic}
		\STATE \textbf{Input}: Observe $\xc_t$ $\rightarrow$ obtain $p(z_{t}| \xc_{t})$
		\STATE Sample $z^{(1)}_t, z^{(2)}_t, \dots, z^{(S)}_t \sim p(z_{t}| \xc_{t})$
		\STATE Count and build $\bm{c}_t$
		\FOR{$r_t=1$ {\bfseries to} $t$}
		\STATE Evaluate $\Psi^{(r)}_t$ using \eqref{predictive}
		\STATE Calculate $p(r_t,\bm{c}_{1:t})$ 
		\STATE Obtain $p(\mathbf{c}_{1:t}) = \sum_{r_t}p(r_t,\mathbf{c}_{1:t})$
		\STATE Compute $p(r_t|\mathbf{c}_{1:t})$
		\STATE Update  $\alpha_{t+1}^{k} = \alpha_{t}^{k} +  \bm{c}_t^{k}\quad\forall k\in\{1,...,K\}$ 
		\ENDFOR
		\STATE \textbf{Return:} $r_t^{\star} =\arg\max
		p(r_t|\mathbf{c}_{1:t})$
	\end{algorithmic}
\end{algorithm}

Notice from the first term of \eqref{likelihood} and the definition of $\bm{c}_t$ that by taking the proposed multinomial model, we are not working with distributions over the $S$-dimensional sampled vectors themselves but over equivalence classes, where two sampled vectors are equivalent $\bm{z}_{S_1}^{\star}\sim\bm{z}_{S_2}^{\star}$ iff their associated counting vectors satisfy $\bm{c}_{S_1}=\bm{c}_{S_2}$. That is, if the vector $\bm{z}_{S_2}^{\star}$ is a permutation of the vector $\bm{z}_{S_1}^{\star}$.  

We now wish to infer the parameter vector $\bm{\theta}_t^{(r)}$ related to the current run-length $r_t$ and its associated data. To carry out the inference method depicted in Figure \ref{fig:thread} we need to find the predictive distribution conditioned on the run length $r_{t-1}$ and the previous data within the referred partition. Marginalizing out the parameters we have
\begin{equation}
p(\bm{c}_t|r_{t-1}, \bm{c}_{1:t-1}^{(r)})= \int p(\bm{c}_t| \bm{\theta}_t) p(\bm{\theta}_t| r_{t-1}, \bm{c}_{1:t-1}^{(r)}) d\bm{\theta}_t,
\end{equation}
where the predictive term $\Psi^{(r)}_t :=p(\bm{c}_t|r_{t-1}, \bm{c}_{1:t-1}^{(r)})$ has not closed form but it is a function of the statistics of the model and its computation is straightforward
$$\Psi^{(r)}_t = \displaystyle\frac{\Gamma(S+1)\Gamma(S_{\alpha})\prod_{k=1}^K\Gamma(c_t^{k}+\alpha_{t-1}^{k})}{\prod_{k=1}^K\Gamma(c_t^{k}+1)\prod_{k=1}^K\Gamma(\alpha_{t-1}^{k})\Gamma(S+S_{\alpha})},$$
where we have defined $S_{\alpha}:=\sum_{k=1}^K\alpha_{t-1}^{k}$. Additionally, using both the binomial coefficient definition and the Gamma function property 
$\Gamma(n+1)=n!$ for $n\in\mathbb{N}$, we transform the previous expression to the following one:
\begin{equation}
\label{predbin}
\Psi^{(r)}_t = \binom{S+S_{\alpha}-1}{S}^{-1}\prod_{k=1}^K\binom{c_t^{k}+\alpha_{t-1}^{k}-1}{c_t^{k}}.
\end{equation}

The term $S_{\alpha}$ grows by $S$ at each time step, leading to numerical instabilities in the l.h.s term of \eqref{predbin} for high values of $t$. Therefore, we have considered the following expression that is numerically more stable and is a result of manipulations on the terms of \eqref{predbin}, it is

\begin{equation}
\label{predictive}
\displaystyle \Psi^{(r)}_t =\prod_{k=1}^{K}\prod_{j=0}^{c_t^{k}-1}\frac{\alpha_{t-1}^k + j}{S_{\alpha} + S_{c}^{(k-1)} + j}\frac{S_{c}^{(k-1)}+j+1}{j+1},
\end{equation}
with $S_{c}^{(k-1)}:= \sum_{l=1}^{k-1}c_t^{l}\quad\forall k=1\dots K$. Notice from the previous expression and the general model equation \eqref{eq:recursive} that the computational cost for a particular time-step grows linearly with $S$. Finally, Algorithm \ref{alg:simplified} presents all steps that must be followed to obtain $r^{\star}_t$ from the initial sequence of observations $\xc_{1:t}$. Notice that $r^{\star}_t$ corresponds to a MAP estimate at each time-step $t$, and it is the variable that we will use to show the most likely CPs in the following experimental results.

\begin{table*}[ht!]
	\vspace*{-0.5\baselineskip}
	\caption{Multinomial CPD vs. Hierarchical CPD metrics. All delay values $(\times 10)$.}
	\centering
	\begin{tabular}{ccccccccc}
		\toprule
		&$S=10$&$S=50$&$S=100$ & \textsc{Hier.}&$S=10$&$S=50$&$S=100$ & \textsc{Hier.}\\
		$\eta$ & \textsc{cpd rate} & \textsc{cpd rate} &\textsc{cpd rate} & \textsc{cpd rate}&\textsc{delay}&\textsc{delay} &\textsc{delay} & \textsc{delay}\\
		\midrule
		$2.0$ & - & $0.12$ & $\bm{0.32}$ & - & $\infty$ & $5.33 \pm 2.30$ & $\bm{5.37} \pm \bm{1.59}$ & $\infty$  \\
		$3.0$& $0.52$ & $\bm{0.88}$ & $0.84$ & $0.2$& $5.30 \pm 2.09$ & $5.68 \pm 3.01$  & $\bm{4.20} \pm \bm{2.17}$ &$10.0 \pm 7.87$  \\
		$4.0$ & $0.88$ & $0.96$ & $\bm{1.0}$ & $0.76$ & $3.57 \pm 2.15$ & $3.28 \pm 2.53$ & $\bm{2.30} \pm \bm{0.96}$ & $5.27+2.00$  \\
		$10.0$ & $0.96$ & $1.0$ & $\bm{1.0}$ & $0.96$& $2.06 \pm 1.77$ & $1.32 \pm 0.39$ & $\bm{1.31} \pm \bm{0.40}$ & $3.52 \pm 2.00$  \\
		\bottomrule
		\label{tab:experiments}
	\end{tabular}
\end{table*}

\section{Experiments}
\label{sec:experiments}

\begin{figure*}[ht!]
	\centering
	\includegraphics[width=1.0\textwidth]{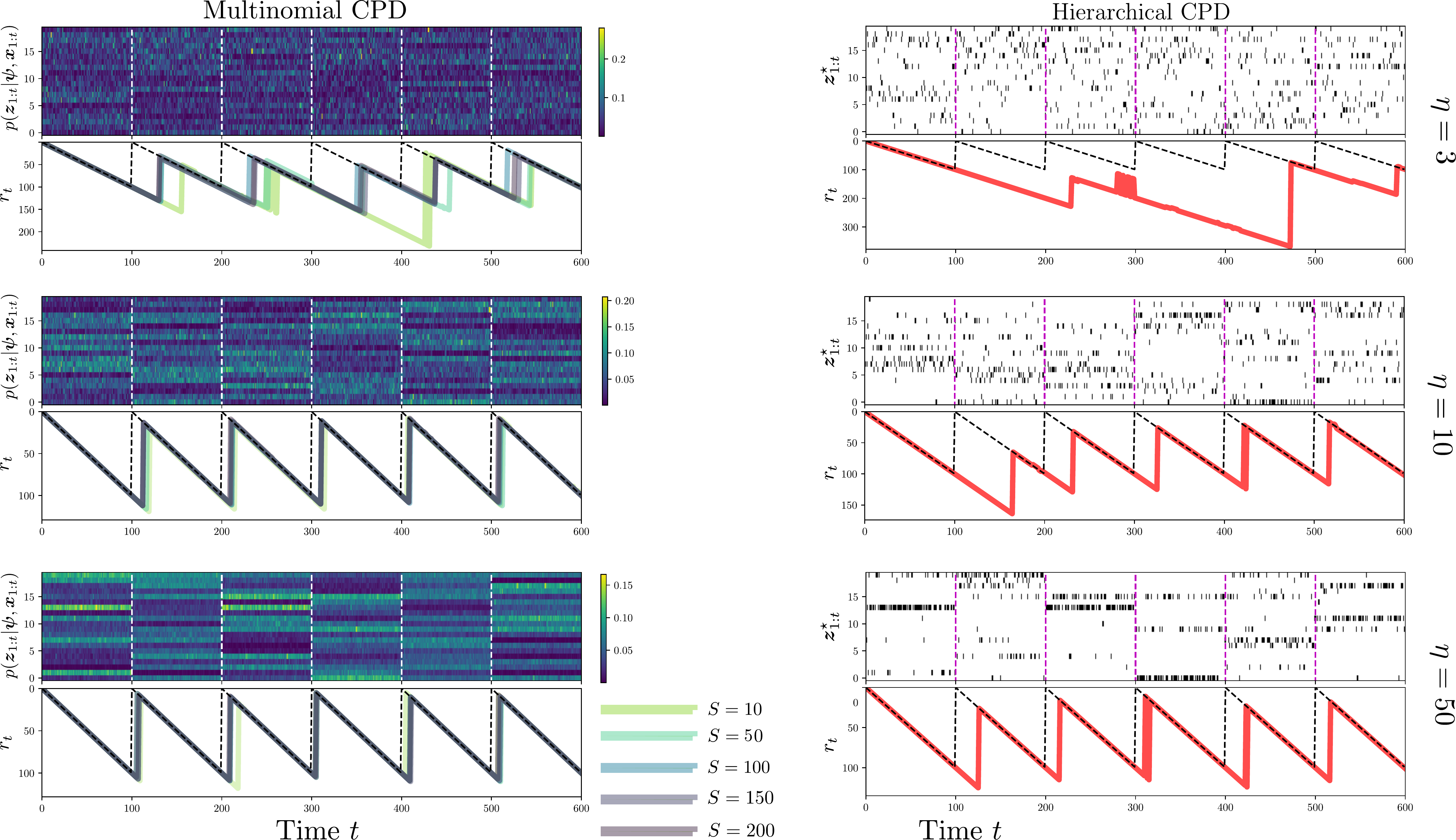}
	\vspace*{-0.5\baselineskip}
	\caption{Comparison between the multinomial CPD, based on sampling from the latent class posterior, and the baseline CPD method. The resulting CPs (bottom figures) are considered as jumps over the MAP estimates (solid lines) of the run-length $r_t\,\forall t$. Dashed lines indicate the true change points. \textbf{Left Column:} Each row represents an example with a more or less flat posterior distribution (upper figures) indicated by $\eta$. Colors of the $r_t$ lines indicate the number of samples $S$ used. \textbf{Right Column:} Results for CPD from different point-estimate pseudo-observations $\zc^{\star}_{1:t}$ (upper figures).}
	\vspace*{-0.5\baselineskip}
	\label{fig:uniform}
\end{figure*}

In this section we evaluate the performance of the proposed multinomial sampling extension for hierarchical CPD. First, we study the improvements of the method (named Multinomial CPD), over synthetic data, where we may increase or decrease the quality of inference over the latent variable to prove that detection is still reliable. In the second experiment, we evaluate the method using real-world data of a monitored user from an authorized human behavior study, analysing how we are able to reduce the delay in the whole detection process. In the experiments, we consider that a change point is detected at time-step $t = t'$ if there is an abrupt decrease from $r^*_{t'-1}$ to $r^*_{t'}$, which means that the CP occurred at instant $t = t'-r^*_{t'}$. We set $r^*_t<r^*_{t-1} - 20$ as the condition for detection.\\ \\
\textbf{5.1. Synthetic data.}~ In our first experiment, the Multinomial CPD model has been applied to sequences of synthetic data and the results have been summarized in Figure \ref{fig:uniform} and Table \ref{tab:experiments}. Particularly, we want to evaluate the performance of the method for several sampling sizes $S$, drawn at each time step and for different levels of \textit{flatness} of the generative posterior distribution.

We have fixed the number of CPs on the latent sequence to five, that is, six partitions, each one occurring every $100$ time steps. Moreover, we have run the algorithm for $T = 600$. In the experiment, the posterior distributions $p(z_{t}| \xc_{t})$ of the latent variables are simulated. For each partition $\rho$, we have generated a set of $100$ $K$-dim vectors $\bm{\theta}_{\rho,t}$ from a Dirichlet distribution with parameters $\bm{\beta}_{\rho}$. At the same time, these $6$ $K$-dim vectors $\bm{\beta}_{\rho}$ have been sampled from a Uniform distribution in the interval $(0,\eta)$. This parameter $\eta$ defines the \textit{flatness} of the synthetic posterior distribution, where a lower $\eta$ implies a flatter generative distribution. The hyperparameter $K$ has been fixed to $20$ classes for the whole experiment. In the proposed model, each $S$-vector has been sampled from a Multinomial($\bm{\theta}_{\rho,t},S$) with the vector $\bm{\theta}_{\rho,t}$ previously presented.

The prior probability of the run length $r_t$ is a function of the hyperparameter $\lambda^{-1}$, which controls the prior probability of a change: the higher is $\lambda$, the less probable is a change. For the Multinomial CPD method (MCPD), we have defined it as a function of the number of samples $\lambda = 10^S$ to make both comparable the terms involved in $\eqref{eq:recursive}$ and also the results in the experiment for different number of samples.  The intuition behind this choice is that, for high values of $S$, we want the prior probability of a change to be almost $0$, so that the change point occurrence is determined from the data. However, more accurate results may be found by tuning the $\lambda$ parameter at each particular case. For the comparison with the Hierarchical CPD method (HCPD) we considered the same values except for the hyperparameter lambda, that has been fixed to $10^{20}$ independently of the \textit{flatness} level of the simulated distributions.

In Figure \ref{fig:uniform} we compare the MCPD (left column) for different number of samples $S = 10,50,100, 150, 200$ with the HCPD (right column) and different levels of \textit{flatness} $\eta = 3.0,10.0,50.0$ (each row). In the upper figures we can see the distributions of the latent variables or the MAP assignments at each time step, respectively. In the bottom figures the MAP estimates of the run-length $r_t$ are jointly shown with dashed lines indicating the true change points. We have also summarized the result of running the method five times for each pair of values ($S$, $\eta$) in table 1. There, we show the average of the \textit{precision}, defined as the ratio of change points detected for each pair, and the mean and standard deviation of the \textit{delay}, defined as the time points between the instant of the detection $t=t'$ and the real instant $t = t'-r^*_{t'}$ in which the CP occurred. For example, if a CP is detected at $t= 150$ and $r^*_{150} = 30$, this means that a change occurred at $t=120$, and the delay of the detection would be $30$ steps.
Looking at the figure results, with the MCPD we detect the five change points for every value of $\eta$ and many of the values of $S$ considered. In the table we confirm that the precision increase as $S$ grows, detecting less change points when the distribution is highly flat for lower values of $S$ and in particular for the HCPD,  that would correspond to the limit case in which $S=1$. For $\eta = 2.0$ no change points are detected using the HCPD method. However, with the MCPD, even if the distribution is that flat we are able to find the change points by increasing the number of samples, obtaining a precision of $88\%$ for $50$ samples at $\eta = 3.0$ versus the $20\%$ in the HCPD case, or even a $100\%$ of precision already at $\eta =4.0$ when $S=100$.

For higher values of $\eta$, we can see both in the Figure \ref{fig:uniform} and the Table \ref{tab:experiments} that the performance is good for both methods in terms of precision. However, the delay of the detections is always notably lower in the proposed MCPD. In comparison to the HCPD, we can see in the table that the average of the delay in the detections is reduced by more than a half when $100$ samples are considered, independently of the \textit{flatness} of the distribution, with just $23.08$ time steps of average delay when $\eta =4.0$ or $13.1$ when $\eta=10.0$.\\ \\
\textbf{5.2. Human behavior.}~ The data are part of a human behavior study with daily measurements obtained by anonymized monitoring of users using their personal smartphones. The monitoring and pre-processing of data was performed by the Evidence-Based Behavior ($\textrm{eB}^2$) \textit{app} between April, 2019 and March, 2020 \cite{Berrouiguet2018}. 

\begin{figure}[ht!]
	\centering
	\includegraphics[width=0.45\textwidth]{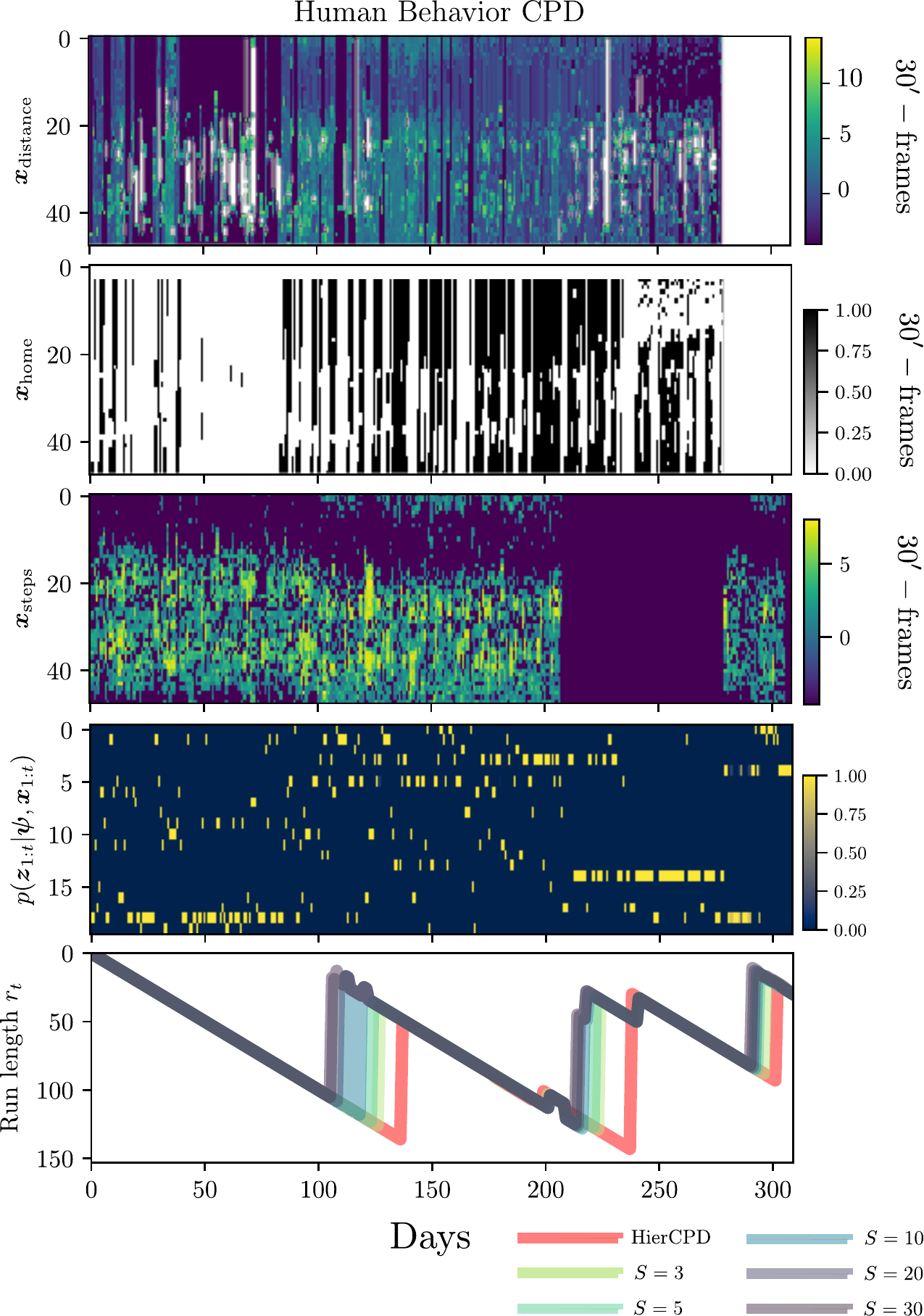}
	\caption{Human behavior CPD with heterogeneous daily mobility metrics from a user.\ \textbf{Three upper rows.} Respectively, 310 days of distance wandered, presence at home and number of steps every 30 minutes.\ \textbf{Fourth row.} Posterior expectations over the $K=20$ latent class indicator $z_t$.\ \textbf{Fifth row.} Hierarchical CPD for several multinomial-sampling cases.}
	\label{fig:behavior}
\end{figure}

From monitored raw traces of latitude-longitude pairs, we calculate distance in kilometers between sequential locations and its global distance to the user starting point, i.e., his/her home. After splitting all data into 30-minutes frames per 24h, we obtained three multivariate heterogeneous observations per day: i) $\xc_{\text{distance}} \in \mathbb{R}^{48}$, ii) $\xc_{\text{home}} \in \{0,1\}^{48}$, where $1$ means staying at home and $0$ otherwise, and iii) $\xc_{\text{steps}} \in  \mathbb{R}^{48}$, where the real-positive values where mapped to real-valued using the mapping $\log(1+y)$. We introduced an \textit{heterogeneous} mixture model given that each daily observation is $\xc_{t} = \{\xc_{\text{distance}}, \xc_{\text{home}}, \xc_{\text{steps}}\}$. We refer to \textit{heterogeneous} as a mix of statistical data types. Additionally, we assume that there is a single latent class indicator $z_t$ that indicates the behavioral profile that the user has followed on that day. The last step is to obtain the complete sequence of posterior estimates $p(\zc_{1:t}| \xc_{1:t})$ via the EM algorithm. The learning method of the mixture model can be adapted to the online nature of CPD using \cite{cappe2009online} or \cite{MorenoRamirezArtes20} if the number of classes $K$ is unbounded. Results obtained are shown in Figure \ref{fig:behavior} for different number of samples drawn by the posterior distribution over the latent variable. We can see that the method finds three change points around day $100$, day $230$ and day $290$, clearly partitioning the time in four behavioral periods between the first and last day of monitoring. These changes have not been contrasted with external information of the user yet, but the results are consistent in terms of number of detections for every value of $S$ considered, and seem to be coherent with the overview of the distributions in the third raw of the figure. Moreover, we can see that increasing the number of samples at each time step, we can reduce the delay in the detection almost $50$ days w.r.t.\ the hierarchical CPD method.

\vspace*{-\baselineskip}
\section{Conclusion}

In this paper, we have presented a novel methodology for improving the Bayesian CPD algorithm of \cite{Adams2007} with latent variable models. Under the assumption that CPs lie in a lower-dimensional manifold, inference is carried out with pseudo-observations based on posterior point-estimates of the latent variables given the data. We introduced a multinomial-sampling method that improves the detection rate and reduces the delay when we treat with high-dimensional sequences of observations. The analytical tractability in the inference is maintained as well as a low computational cost. The experimental results show significant improvements in the CPD as posterior estimates become less certain. Interestingly, even under a good inference performance, the multinomial sampling method reduces the delay of detection, what in practice is a key point for its application to real-world problems. We illustrate an example on a human behavioral study, that detects changes in the circadian patterns of a user. 
In future work, this could be integrated with other CPD methods that consider the dimensionality of the latent variables unbounded \cite{MorenoRamirezArtes20}.


\bibliographystyle{IEEEbib}
\bibliography{mlsp_2020}

\end{document}